\documentclass[conference]{IEEEtran}
\IEEEoverridecommandlockouts

\usepackage{cite}
\bstctlcite{IEEEexample:BSTcontrol}
\usepackage{amsmath,amssymb,amsfonts}
\usepackage{algorithmic}
\usepackage{graphicx}
\usepackage{textcomp}
\usepackage{xcolor}
\usepackage{flushend}
\usepackage{hyperref}

\usepackage{verbatim}
\usepackage{comment}
\usepackage{float}
\usepackage{amsmath}
\usepackage{booktabs}
\usepackage{multirow}
\usepackage{url}
\usepackage[section]{placeins}

\def\BibTeX{{\rm B\kern-.05em{\sc i\kern-.025em b}\kern-.08em
    T\kern-.1667em\lower.7ex\hbox{E}\kern-.125emX}}

\begin{document}

\title{Do LLMs Surpass Encoders for Biomedical NER?
}

\author{\IEEEauthorblockN{Motasem S Obeidat}
\IEEEauthorblockA{\textit{Department of Computer Science} \\
\textit{University of Kentucky}\\
Lexington, KY USA \\
obeidat.s.motasem@uky.edu}
\and
\IEEEauthorblockN{Md Sultan Al Nahian}
\IEEEauthorblockA{\textit{Division of Biomedical Informatics} \\
\textit{University of Kentucky}\\
Lexington, KY USA \\
mna245@uky.edu}
\and
\IEEEauthorblockN{Ramakanth Kavuluru}
\IEEEauthorblockA{\textit{Division of Biomedical Informatics} \\
\textit{University of Kentucky}\\
Lexington, KY USA \\
ramakanth.kavuluru@uky.edu}
}

\maketitle

\begin{abstract}
Recognizing spans of biomedical concepts and their types (e.g., drug or gene) in free text, often called biomedical named entity recognition (NER), is a basic component of information extraction (IE) pipelines. Without a strong NER component, other applications, such as knowledge discovery and information retrieval, are not practical. State-of-the-art in NER shifted from traditional ML models to deep neural networks with transformer-based encoder models (e.g., BERT) emerging as the current standard. However, decoder models (also called large language models or LLMs) are gaining traction in IE. But LLM-driven NER often ignores positional information due to the generative nature of decoder models. Furthermore, they are computationally very expensive (both in inference time and hardware needs). Hence, it is worth exploring if they actually excel at biomedical NER and assess any associated trade-offs (performance vs efficiency). This is exactly what we do in this effort employing the same BIO entity tagging scheme (that retains positional information) using five different datasets with varying proportions of longer entities. Our results show that the LLMs chosen (Mistral and Llama: 8B range) often outperform best encoder models (BERT-(un)cased, BiomedBERT, and DeBERTav3: 300M range) by 2--8\% in F-scores except for one dataset, where they equal encoder performance. This gain is more prominent among longer entities of length $\geq 3$ tokens. However, LLMs are one to two orders of magnitude more expensive at inference time and may need cost prohibitive hardware. Thus, when performance differences are small or real time user feedback is needed, encoder models might still be more suitable than LLMs.

\end{abstract}

\begin{IEEEkeywords}
named entity recognition, encoder models, large language models
\end{IEEEkeywords}

\section{Introduction}

Biomedical information extraction (IE) is a key natural language processing (NLP) task involving multiple subtasks including named entity recognition (NER), entity normalization, and relation extraction (RE). IE is the backbone of creation of knowledge bases that drive applications in biomedical knowledge discovery. 
NER is often the first and critical step in an IE pipeline, considering how errors in it will snowball to lead to more downstream errors. Specialized NER models have been developed to identify popular entities such as medications~\cite{kim2020ensemble}, diseases~\cite{leaman2013dnorm}, genes~\cite{wei2023gnorm2}, and even specialized concepts such as adverse effects and phenotypes~\cite{10.1371/journal.pdig.0000152, YANG2024100887}.

Over the past decade, deep neural networks have dominated the methods landscape for biomedical NER~\cite{song2021deep, noh2021joint,yadav2018survey} and the current crop of strong baselines come from the transformer architecture~\cite{10.1093/bioinformatics/btz682}, specifically based on the encoder models such as BERT~\cite{devlin2019bert}, RoBERTa, and DeBERTa~\cite{hedebertav3}. These models typically process an input sequence and assign an entity tag per token. This tagging scheme is often referred to as BIO where the \textit{outside} O tag is typically used to mark non-entity tokens, while the B and I tags capture the \textit{begin} and \textit{inside} tokens of an entity, respectively, with the latter reserved for non-singleton entities. More time-consuming span-based NER methods have also risen where contiguous text spans up to a maximum length (say, five tokens) are predicted to form an entity or not~\cite{wadden2019entity, ai2023end}. 

With the advent of OpenAI's GPT-3~\cite{brown2020language}, generative transformers (based on the decoder component of transformers) showed that autoregressive next-token prediction can be a powerful paradigm, with use beyond generative tasks. For example, using instruction finetuning~\cite{weifinetuned, rohanian2024exploring,keloth2024advancing}, many conventionally non-generative tasks can be converted into generative input/output templates. This has been the  strategy to have decoder models, well known as large language models (LLMs), follow instructions and perform IE tasks. This enabled LLMs to act as chat bots such as ChatGPT and Claude. Leveraging  vast amounts of knowledge ingested from massive corpora coupled with instruction finetuning on thousands of datasets, LLMs have become formidable tools in modern NLP, especially in the low training data regime~\cite{hsieh2023distilling, zhouuniversalner}. 

Despite all the hype and excitement surrounding LLMs, we are not aware of any substantial efforts that evaluate their potential for biomedical NER, especially in the supervised setting. Some studies show that LLMs excel in zero-shot and few-shot settings, with the latter involving in-context learning (ICL) where examples are provided as part of the prompt~\cite{monajatipoor2024llms}. However, when full training datasets are used, encoder models are still shown to be better~\cite{hu2024improving, mullick2024intent}. A recent effort based on the Llama2 LLM~\cite{keloth2024advancing} shows some improvements in NER over encoders. However, it considered only three entity types and is more focused on assessing instruction finetuning and generalizability; the improvements are also not consistent and in the 1\% range.  An additional complication in evaluating LLMs for NER is the potential loss of positional information if only answer spans are output as opposed to exact locations of those spans. Although generating only answer spans is more efficient from a generative angle, encoder models naturally output per-token BIO tags, thus providing exact locational information useful for downstream applications. Outputting answer spans (just text strings) can lead to loss of repeated entity mentions and can adversely effect downstream RE components. Without exercising care in choosing appropriate output templates that preserve entity locations, evaluations can become apples-to-oranges comparisons.  

 Another factor influencing NER model performance is how well longer entities are recognized. Previous studies have indicated that longer and more descriptive entities often present challenges in segmentation and recognition, particularly in biomedical contexts where such entities are prevalent \cite{fu2020rethinking, hong2020dtranner}. In fact, recent research shows that
 NER performance can be inconsistent for datasets containing longer entities, often leading to increased error rates \cite{jeong2021regularization}.

Considering these factors, in this paper, we report on an elaborate effort comparing three encoder models (BERT, BiomedBERT~\cite{gu2021domain}, and DeBERTa) and two LLMs (Mistral~\cite{jiang2023mistral7b} and Llama~\cite{grattafiori2024llama}) with identical BIO tagging output scheme across five different biomedical NER datasets with varying proportions of longer entities. We also report on average inference times per each test instance. The data and code used in our effort are made available here: \url{https://github.com/bionlproc/LLMs-vs-Encoders-for-BioNER}

\section{Methodology}


\subsection{Datasets}
We use five publicly available biomedical datasets: JNLPBA, BioRED, ChemProt, BC5CDR, and Reddit-Impacts. They contain annotations of entities such as diseases, proteins, and chemicals. These datasets contain varying numbers of longer entities, which for the purposes of this paper are those with length $\geq 3$, ranging from 10\% to over 50\% of the full test datasets. 

Based on the GENIA corpus, the JNLPBA dataset was used in a collaborative task for biomedical entity recognition. It comprises 2,000 MEDLINE abstracts annotated with five key biomedical entity categories: proteins, DNA, RNA, cell lines, and cell types. The objective was to assess systems proficient in identifying these elements inside complex biomedical texts, establishing it as a baseline for NER systems focused on molecular biology~\cite{collier2004introduction}.
BioRED is a relatively new, large biomedical data set for NER and RE tasks. It consists of 600 PubMed abstracts annotated with six entity types: genes, diseases, chemicals, variants, species, and cell lines. BioRED was developed to ensure diversity in the types of  entities involved, challenging models to deal with multiple entity types simultaneously~\cite{luo2022biored}.

Started in 2004~\cite{hirschman2005overview}, the BioCreative challenge series represents a collaborative effort to assess text-mining and information extraction systems for the biomedical domain with emphasis on NER and RE tasks. The BioCreative VI challenge~\cite{10.1093/database/bay147} introduced the ChemProt dataset to evaluate models that extract relationships between chemical compounds and proteins (gene products). It is also extensively used for NER evaluation focusing on chemical and gene/protein entities. The dataset has 4,966 PubMed abstracts annotated with chemical and gene/protein entities. ChemProt is notorious for complex chemical names involving a mix of alphabetical, numerical, and special characters. We used the ChemProt dataset as tokenized and made available by~\cite{ai2023end}.
BC5CDR~\cite{li2016biocreative} is a well-established dataset created for the BioCreative V challenge. The dataset consists of 1,500 PubMed abstracts with 4,409 chemical entities and 5,818 diseases, in addition to 3,116 chemical-disease interactions. 

\begin{table}[H]
  \centering
  \renewcommand{\arraystretch}{1.3}
  \caption{Instance Stats Across Training, Validation, and Test Sets}
  \label{Datasets}
  \begin{tabular}{lrrr}
    \hline
    \textbf{Dataset} & \textbf{Train} & \textbf{Dev} & \textbf{Test} \\
    \hline
    \text{JNLPBA} & 16,807 & 1,739 & 3,856 \\
    \text{BioRED} & 1,051 & 269 & 273 \\
    \text{ChemProt}  & 13,305 & 3,360 & 8,179 \\
    \text{BC5CDR} & 4,560 & 4,581 & 4,797 \\
    \text{Reddit-Impacts} & 842 & 258 & 278 \\
    \hline
  \end{tabular}
\end{table}

The Reddit-Impacts dataset contains Reddit posts annotated for clinical and social impacts of substance use disorders. It has 1,380 posts discussing non-medical use of substances, particularly opioids, stimulants, and benzodiazepines. Roughly 23\% of the total posts contain words or phrases annotated as clinical or social impacts, with 246 posts containing a clinical impact tag and 72 posts with a social impact tag. This is the only social media dataset considered for this paper and has many longer entities~\cite{ge2024reddit}.  
As shown in Table~\ref{Datasets}, as per the distribution of instances across training, validation, and test sets, Reddit-Impacts is the smallest dataset. However, the difference in size between Reddit-Impacts and BioRED (the second smallest) is relatively minor compared to the much larger gap observed between the sizes of JNLPBA and BC5CDR (the second and third largest datasets). This highlights that while Reddit-Impacts is the smallest, the variation in dataset sizes among the smaller datasets is less pronounced than the differences among the larger ones.

\begin{table}[h]
  \centering
  \footnotesize 
  \setlength{\tabcolsep}{3pt}
  \renewcommand{\arraystretch}{1.5}
  \caption{Entity Distribution by Length Across Test Datasets}
  \label{tbl:Entites_Count}
  \begin{tabular}{lrrrr}
    \hline
    \textbf{Dataset} & \textbf{Total} & \textbf{Length 1} & \textbf{Length 2} & \textbf{Length $\geq 3$} \\
    \hline
    \text{JNLPBA} & 8,662 & 3,466 (40.01\%) & 2,620 (30.24\%) & 2,576 (29.73\%) \\
    \text{BioRED} & 3,503 & 2,525 (72.08\%) & 411 (11.73\%) & 567 (16.18\%) \\
    \text{ChemProt} & 2,984 & 1,348 (45.17\%) & 613 (20.54\%) & 1,023 (34.28\%) \\
    \text{BC5CDR} & 9,809 & 7,407 (75.51\%) & 1,438 (14.66\%) & 964 (9.82\%) \\
    \text{Reddit-Impacts} & 80 & 28 (35.0\%) & 7 (8.75\%) & 45 (56.25\%) \\
    \hline
  \end{tabular}
\end{table}

Table~\ref{tbl:Entites_Count} shows total number of entities, along with proportions of test entities of lengths one (e.g., ``Parathormone''), two (e.g., ``T lymphocytes''), and $\geq 3$ (e.g., ``peripheral blood lymphocyes'') across datasets. BC5CDR and BioRED report a high proportion ($\geq 70\%$) of single-token entities. On the other hand, JNLPBA and ChemProt feature a relatively equal spread on the measures, with the single-token entities making up 40.01\% and 45.17\% of the totals, respectively. 
Reddit-Impacts, has a unique distribution where over half of the entities are of length $\geq 3$. After this dataset, ChemProt has the highest proportion of length $\geq 3$ entities constituting a third of that dataset. Here it is important to specify that length is calculated by the number of tokens in the BIO tagged gold datasets as opposed to the tokens used by the vocabularies of different language models. This ensures consistent evaluation across different models, where length assessment is decoupled from how encoders and decoders are using different subword tokens in their vocabularies.

\vspace{-1mm}
\subsection{Model Selection}
The central question we want to answer in this paper is whether LLMs excel over encoder models when full training datasets are available. As outlined in the introduction, by employing the BIO tagging scheme, we ensure that our models produce outputs that are both precise and compatible with common evaluation metrics in the field~\cite{banerjee2021biomedical}.

Encoders use a bidirectional attention mechanism in order to attend to the previous and the next tokens simultaneously. This makes these models particularly suitable for language understanding tasks that rely heavily on the context surrounding each token, which is critical for token classification tasks such as NER~\cite{devlin2019bert}. For our experiments, we selected BERT-large-uncased (336M parameters), BERT-large-cased (336M parameters), BiomedBERT-large-uncased-abstract (336M parameters), and DeBERTa-v3-large (435M parameters) models; we chose the `large' variants  to make fair comparisons with the even larger decoder-based models used in this work. Please note that BiomedBERT was earlier called PubmedBERT and is unique from other biomedical encoders considering it is the first model to be pre-trained from scratch using biomedical abstracts with a custom vocabulary derived from PubMed abstracts. This is in contrast with other models such as BioBERT~\cite{10.1093/bioinformatics/btz682} that uses continuous pre-training on top of a general domain model with the latter's vocabulary. Also note that BiomedBERT does not have a cased variant and DeBERTa does not have an uncased variant. 

\begin{figure}[ht!]
    \centering
\includegraphics[width=1\linewidth]{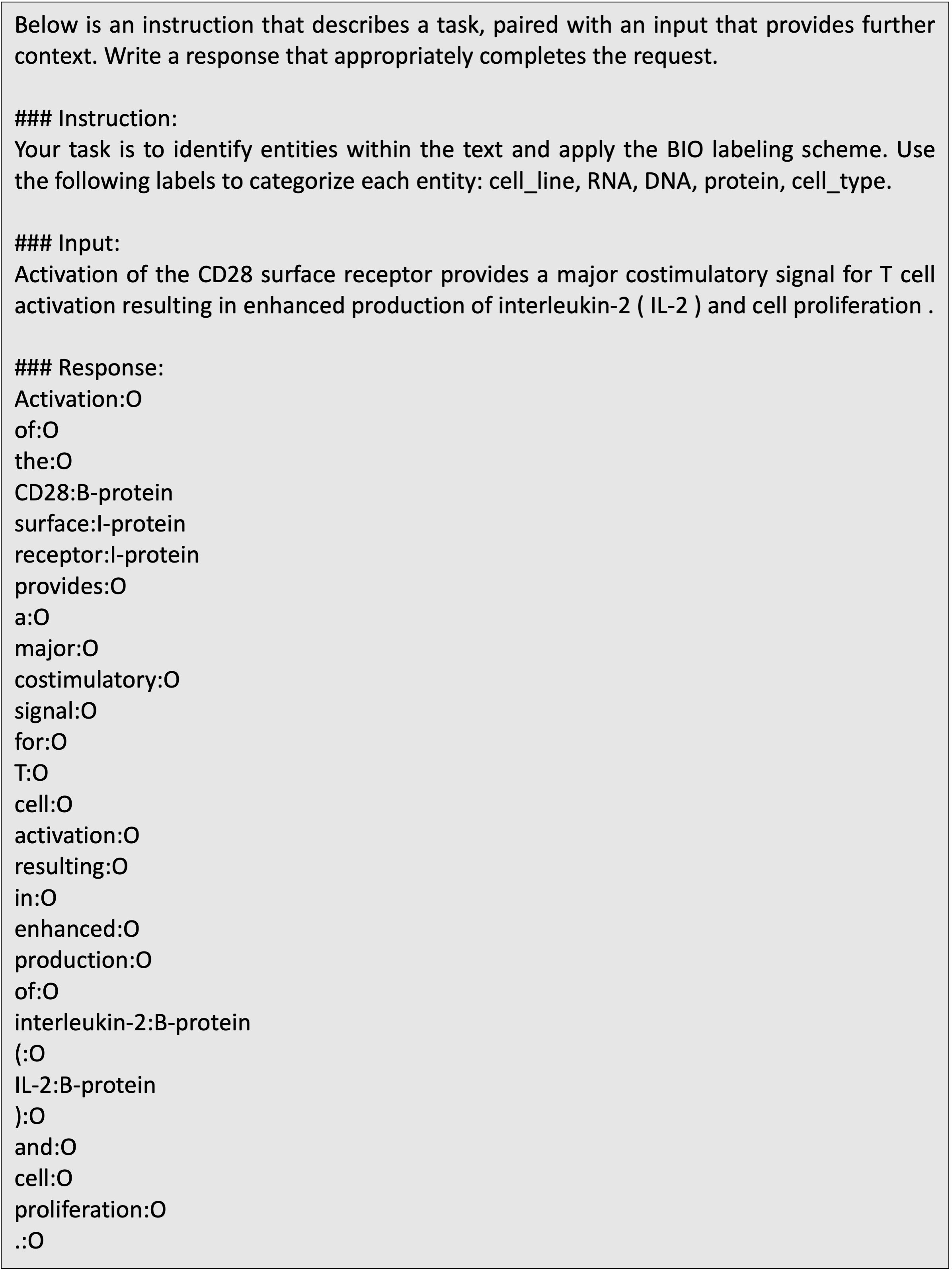}
    \caption{Sample prompt for the JNLPBA dataset for LLM driven NER}
    \label{fig:Prompt}
\end{figure}

A decoder, in turn, employs autoregressive methods to predict the next token in a sequence, making it well-suited for generative tasks \cite{krishnan2024towards}. However, as we already broached earlier, the ``instruct'' variants of modern LLMs are finetuned for non-generative tasks also, via appropriate prompts. So we frame NER as a prompt-driven task, making the models produce outputs using the BIO scheme. We use Mistral-7B-Instruct-v0.3 (7B parameters) and Llama-3.1-8B-Instruct (8B parameters), both of which are publicly available. Both models were further optimized using a 4-bit quantized QLoRA configuration~\cite{dettmers2024qlora} with LoRA (low-rank adaptation) settings of $r=128$, $\alpha=256$, and a dropout of $0.05$, making Mistral's effective trainable size 341M parameters with QLoRA adaptation; Llama is likewise used with 353M trainable parameters. With carefully designed prompts, we leverage the strengths of instruction finetuned decoder-based models to obtain entity tags required for NER. Figure \ref{fig:Prompt} provides an example prompt using a sample from the JNLPBA dataset.

\subsection{Experimental Setup}
\begin{table*}[ht!]
\centering
\caption{Strict precision, recall, and F1 Scores for test sets across datasets and models \\ (For LLMs, the parameter size in the 2nd column is that of the finetuned QLoRA component)}
\label{tbl:Results}
\renewcommand{\arraystretch}{1.3}
\resizebox{\textwidth}{!}{%
\begin{tabular}{l l c c c c c c c c c c c c c c}
\toprule
\multirow{2}{*}{\textbf{Dataset}} & \multirow{2}{*}{\textbf{Model Type}}  & \multicolumn{3}{c}{\textbf{Overall Test}} & \multicolumn{3}{c}{\textbf{Test Entity Length 1}} & \multicolumn{3}{c}{\textbf{Test Entity Length 2}} & \multicolumn{3}{c}{\textbf{Test Entity Length 3+}} \\
\cmidrule(lr){3-5} \cmidrule(lr){6-8} \cmidrule(lr){9-11} \cmidrule(lr){12-14} 
 &  & \textbf{P} & \textbf{R} & \textbf{F1} & \textbf{P} & \textbf{R} & \textbf{F1} & \textbf{P} & \textbf{R} & \textbf{F1} & \textbf{P} & \textbf{R} & \textbf{F1} \\
\midrule
\multirow{4}{*}{\parbox{2.25cm}{JNLPBA \\ \footnotesize \textit{Length 3+: 29.73\%}}} 
 & BERT (large-uncased) (336M)
 & 67.77 & 74.29 & 70.88 & 50.22 & 82.03 & 62.30 & 87.80 & 81.30 & 84.42 & 65.64 & 69.88 & 67.69 \\
 & BERT (large-cased) (336M)
 & 67.71 & 75.78 & 71.52 & 48.79 & 83.53 & 61.60 & 88.05 & 82.98 & 85.44 & 65.89 & 70.88 & 68.29 \\
 & BiomedBERT (large-uncased) (336M)
 & 69.70 & 76.83 & 73.09 & 51.53 & 84.74 & 64.08 & 88.91 & 82.33 & 85.49 & 67.82 & 72.95 & 70.29 \\
 & DeBERTa (v3-large) (435M)
 & 69.88 & 77.70 & 73.59 & 51.05 & 86.56 & 64.22 & 88.28 & 82.79 & 85.44 & 66.65 & 71.26 & 68.88 \\
 & Mistral (7B-Instruct-v0.3) (341M)
 & 72.18 & 79.52 & \textbf{75.67} & 53.02 & 88.32 & \textbf{66.26} & 91.67 & 85.65 & \textbf{88.56} & 68.78 & 72.78 & 70.72 \\
 & Llama (3.1-8B-Instruct) (353M)
 & 71.11 & 78.99 & 74.85 & 49.43 & 88.08 & 63.32 & 91.51 & 85.19 & 88.24 & 70.18 & 73.84 & \textbf{71.96} \\
 \midrule
\multirow{4}{*}{\parbox{2.25cm}{BioRED \\ \footnotesize \textit{Length 3+: 16.18\%}}}
 & BERT (large-uncased) (336M)
 & 82.05 & 85.37 & 83.68 & 85.26 & 92.79 & 88.87 & 77.49 & 78.42 & 77.95 & 61.89 & 69.28 & 65.38 \\
 & BERT (large-cased) (336M)
 & 81.22 & 86.89 & 83.96 & 82.34 & 94.02 & 87.80 & 81.92 & 83.69 & 82.80 & 61.67 & 72.16 & 66.51 \\
 & BiomedBERT (large-uncased) (336M)
 & 87.33 & 90.90 & \textbf{89.08} & 88.09 & 96.08 & 91.91 & 85.99 & 86.81 & 86.40 & 72.23 & 80.38 & 76.09 \\
 & DeBERTa (v3-large) (435M)
 & 87.57 & 90.23 & 88.88 & 88.05 & 95.09 & 91.43 & 88.24 & 86.33 & \textbf{87.27} & 75.00 & 79.41 & 77.14 \\
 & Mistral (7B-Instruct-v0.3) (341M)
 & 88.81 & 88.12 & 88.46 & 84.96 & 94.18 & 89.33 & 85.18 & 81.29 & 83.19 & 80.53 & 80.67 & \textbf{80.60} \\
 & Llama (3.1-8B-Instruct) (353M)
 & 88.94 & 88.21 & 88.58 & 88.86 & 96.04 & \textbf{92.31} & 83.93 & 78.90 & 81.33 & 76.35 & 75.95 & 76.15 \\
 \midrule
\multirow{4}{*}{\parbox{2.25cm}{ChemProt \\ \footnotesize \textit{Length 3+: 34.28\%}}}
 & BERT (large-uncased) (336M)
 & 49.61 & 37.97 & 43.02 & 27.22 & 48.22 & 34.80 & 83.47 & 45.02 & 58.49 & 48.10 & 34.08 & 39.90 \\
 & BERT (large-cased) (336M)
 & 54.49 & 36.66 & 43.83 & 35.77 & 45.33 & 39.99 & 72.53 & 41.09 & 52.46 & 46.78 & 31.03 & 37.31 \\
 & BiomedBERT (large-uncased) (336M)
 & 68.41 & 40.32 & 50.74 & 38.82 & 47.26 & 42.62 & 85.80 & 45.62 & 59.57 & 63.38 & 37.87 & 47.41 \\
 & DeBERTa (v3-large) (435M)
 & 50.40 & 41.59 & 45.57 & 31.33 & 50.59 & 38.70 & 79.74 & 45.77 & 58.16 & 50.93 & 38.29 & 43.72 \\
 & Mistral (7B-Instruct-v0.3) (341M)
 & 73.74 & 40.68 & 52.43 & 53.99 & 48.66 & \textbf{51.19} & 83.00 & 44.26 & 57.73 & 70.12 & 35.72 & 47.33 \\
 & Llama (3.1-8B-Instruct) (353M)
 & 71.73 & 42.06 & \textbf{53.03} & 47.65 & 48.89 & 48.26 & 82.77 & 47.89 & \textbf{60.67} & 74.24 & 39.07 & \textbf{51.20} \\
 \midrule
\multirow{4}{*}{\parbox{2.25cm}{BC5CDR \\ \footnotesize \textit{Length 3+: 9.82\%}}}
 & BERT (large-uncased) (336M)
 & 83.10 & 87.26 & 85.13 & 75.52 & 94.03 & 83.76 & 91.55 & 88.11 & 89.79 & 76.05 & 75.66 & 75.86 \\
 & BERT (large-cased) (336M)
 & 85.56 & 88.59 & 87.05 & 79.98 & 94.83 & 86.77 & 88.55 & 84.98 & 86.73 & 79.22 & 79.38 & 79.30 \\
 & BiomedBERT (large-uncased) (336M)
 & 85.88 & 91.10 & 88.42 & 78.39 & 96.46 & 86.49 & 91.24 & 89.85 & 90.54 & 84.27 & 85.14 & \textbf{84.70} \\
 & DeBERTa (v3-large) (435M)
 & 87.06 & 90.73 & 88.86 & 80.34 & 95.94 & 87.45 & 92.87 & 89.71 & \textbf{91.26} & 84.19 & 82.80 & 83.49 \\
 & Mistral (7B-Instruct-v0.3) (341M)
 & 90.42 & 91.05 & \textbf{90.73} & 84.42 & 96.45 & 90.04 & 93.91 & 87.97 & 90.84 & 83.60 & 79.88 & 81.70 \\
 & Llama (3.1-8B-Instruct) (353M)
 & 89.52 & 89.44 & 89.48 & 84.88 & 95.88 & \textbf{90.05} & 91.71 & 86.16 & 88.85 & 81.31 & 76.01 & 78.57 \\
 \midrule
\multirow{4}{*}{\parbox{2.25cm}{Reddit-Impacts \\ \footnotesize \textit{Length 3+: 56.25\%}}}
 & BERT (large-uncased) (336M)
 & 15.14 & 31.11 & 20.36 & 09.09 & 89.29 & 16.50 & 85.71 & 85.71 & \textbf{85.71} & 16.42 & 20.00 & 18.03 \\
 & BERT (large-cased) (336M)
 & 17.46 & 26.19 & 20.95 & 16.81 & 71.43 & 27.21 & 50.00 & 57.14 & 53.33 & 08.33 & 10.20 & 09.17 \\
 & BiomedBERT (large-uncased) (336M)
 & 21.57 & 26.51 & 23.78 & 19.23 & 71.43 & 30.30 & 66.67 & 57.14 & 61.54 & 06.98 & 06.25 & 06.59 \\
 & DeBERTa (v3-large) (435M)
 & 18.49 & 25.29 & 21.36 & 15.60 & 78.57 & 26.04 & 66.67 & 57.14 & 61.54 & 15.79 & 17.31 & 16.51 \\
 & Mistral (7B-Instruct-v0.3) (341M)
 & 37.29 & 27.50 & \textbf{31.65} & 25.32 & 71.43 & \textbf{37.38} & 100.00 & 71.43 & 83.33 & 23.81 & 11.11 & 15.15 \\
 & Llama (3.1-8B-Instruct) (353M)
 & 30.67 & 28.75 & 29.68 & 05.04 & 64.29 & 09.35 & 83.33 & 71.43 & 76.92 & 62.50 & 33.33 & \textbf{43.48} \\
\bottomrule
\end{tabular}%
}
\end{table*}

\subsubsection{Hardware and computational resources}
The experiments were conducted using an NVIDIA H100 GPU with 80GB VRAM. Encoder models (BERT-(un)cased, BiomedBERT, and DeBERTa) were trained and evaluated using a single GPU, while decoder-based models (Mistral and Llama) required two GPUs for both training and evaluation due to their higher computational and memory demands.

\subsubsection{Model training}
For each model, we performed hyperparameter tuning to improve its performance, including changing certain parameters such as learning rates and batch sizes.
For encoder models, a maximum input length of 512 tokens was used, with batch sizes ranging from 4 to 32; learning rates of 1e-5, 2e-5, and 3e-5 were employed during the training for a total of 20 epochs with early stopping criteria of three epochs.
For LLMs (Mistral and Llama), the maximum input length was adjusted dynamically based on the longest sequences across both the training and validation datasets. However, it was capped at 2048 tokens to ensure compatibility and efficiency. The training was done with a fixed batch size of 32, and learning rates of 2e-5 and 4e-5 were employed during the training, for a total of 20 epochs with early stopping criteria of three epochs.

\subsubsection{Performance measures}
Model performance is evaluated following the International Workshop on Semantic Evaluation (SemEval) guidelines\footnote{\href{https://www.davidsbatista.net/blog/2018/05/09/Named_Entity_Evaluation}{https://www.davidsbatista.net/blog/2018/05/09/Named\_Entity\_Evaluation}}. Precision, recall, and F1 score measures are calculated using the entity-level \textit{strict} measure, requiring exact matches for multi-token entities. For an entity to be predicted accurately, it must possess both the correct boundaries and the true entity type. Besides strict evaluation, we also compute entity-level \textit{relaxed} precision, recall, and F1 scores, which allow for partial boundary matches over the surface string, regardless of the entity type. This provides a more flexible assessment of model performance. 

\subsubsection{Tokenization consistency}
Since different models tokenize text differently, we ensured that all predicted entities are mapped back to the original dataset's tokenization scheme before evaluation. This step prevented inconsistencies in entity length and ensured that all models were evaluated on the same tokenized text, making comparisons fair and reliable.

\section{Results and Discussion}


 The scores across all datasets for all models are shown in Table \ref{tbl:Results}. (Unless otherwise specified, the scores discussed in the rest of this section are F-scores.) At a high level, these results suggest that LLMs outperform encoders  in overall F-scores and also for \textit{longer entities} (here means length $\geq 3$) with a couple of exceptions: for BioRED, DeBERTa and BiomedBERT was slightly better than LLMs in the overall F-score. And for BC5CDR, DeBERTa and BiomedBERT had better F-score even for longer entities. While encoders do well on shorter entities, LLMs show relative robustness in handling more complex multi-token entities. Considering precision and recall scores, encoders seems to maintain almost the same recall as LLMs and appear to lose mostly on precision. Mistral seems slightly better than Llama between the two LLMs. Among encoders, the relative advantage of the more sophisticated training approach of DeBERTa vanishes against BiomedBERT --- domain specific pretraining with custom vocabulary wins here over advanced pre-training methods. BiomedBERT, a smaller and faster encoder, beats DeBERTa for BioRED, Reddit-Impacts, and ChemProt datasets and almost has the same score for JNLPBA and BC5DCR; even for these two datasets, it has better scores for longer entities.

 \begin{table}[ht!]
\centering
\caption{Average Inference Time per Test Sample (in seconds)}
\label{tbl:Inference_AvgTimes}
\renewcommand{\arraystretch}{1.2}
\begin{tabular}{l l c}
\toprule
\textbf{Dataset} & \textbf{Model} & \textbf{Avg Inf.~Time } \\
\midrule
\multirow{4}{*}{\parbox{1.5cm}{JNLPBA}}
& BERT-Family (Avg)  & 0.0274 \\           
    & DeBERTa-v3-Large  & 0.0493 \\
    & Mistral-7B  & 2.0897 \\
    & Llama-8B  & 1.7097 \\
\midrule
\multirow{4}{*}{\parbox{1.5cm}{BioRED}}
& BERT-Family (Avg) & 0.0441 \\          
    & DeBERTa-v3-Large  & 0.0669 \\
    & Mistral-7B & 11.5464 \\
    & Llama-8B   & 9.6395 \\
\midrule
\multirow{4}{*}{\parbox{1.5cm}{ChemProt}}
& BERT-Family (Avg) & 0.0265 \\             
    & DeBERTa-v3-Large & 0.0493 \\
    & Mistral-7B  & 2.2937 \\
    & Llama-8B  & 2.1018 \\
\midrule
\multirow{4}{*}{\parbox{1.5cm}{BC5CDR}}
& BERT-Family (Avg) & 0.0260 \\           
    & DeBERTa-v3-Large  & 0.0477 \\
    & Mistral-7B &  2.0953 \\
    & Llama-8B  & 1.8434 \\
\midrule
\multirow{4}{*}{\parbox{1.5cm}{Reddit-Impacts}}
& BERT-Family (Avg)  & 0.0268 \\                   
    & DeBERTa-v3-Large & 0.0496 \\
    & Mistral-7B & 1.9446 \\
    & Llama-8B   & 1.8815 \\
\bottomrule
\end{tabular}%
\end{table}

Coming to individual datasets, for JNLPBA where 40.01\% of entities are of length one, 30.24\% being length two, and 29.73\% being longer, the gap between DeBERTa/BiomedBERT and LLMs is similar across all length groups. LLMs were more tolerant of variations in the length of the entities, with Mistral getting 75.67\% overall and 70.72\% for longer entities. Strangely, LLMs had better F-scores (by $\approx 4$ points) for longer entities than singleton mentions, mostly due to high losses in precision in the latter group. Among encoders, while DeBERTa wins overall, BiomedBERT is better for longer entities.

BioRED has the highest percentage of singletons at 72.08\%, which may not have left much scope for models that excel on longer entities. BiomedBERT and DeBERTa did well on singletons, getting F-scores 91.91\% and 91.43\%, respectively; their performance dropped steeply by nearly 15 points for longer entities. LLMs performed more consistently across lengths. Especially, Mistral ended with the smallest performance gap across lengths, with 88.46\% overall F-score and 80.60\% for longer entities. However, the overall winner for this dataset is BiomedBERT by a very slim margin. 

Relative to BioRED, the ChemProt dataset  has a more balanced distribution of entity lengths. This is the only dataset where Llama was better than Mistral (overall and also for longer entities). It is also peculiar to see that, except for Mistral and BERT-cased, the longer entity F-scores were higher than singleton scores, again, owing to major precision issues for one-token entities. 
Like BioRED, the BC5CDR dataset also suffers from over-saturation with singleton entities at 75.51\%. For both encoders and LLMs, scores dipped for longer entities compared to singletons. 

Reddit-Impacts is the only dataset with majority longer entities (56.25\%). It has the widest margin of nearly eight F1 points between LLMs and encoders. This gap is over 20\% for longer entities. 
That said, even the top performance (by Mistral) in this dataset is only 31.65\%, which indicates that complex long entities are hard for current models, LLMs or encoders. Some of this is tempered by the fact that the entire test set has only 80 entities and hence the small sample size may not lead to highly reliable trends here. 


\begin{table}[ht!]
\centering
\caption{Relaxed precision, recall, and F1 Scores for test sets across datasets and models}
\label{tbl:Relaxed_Precision_Recall_F1}
\renewcommand{\arraystretch}{1.2}
\begin{tabular}{l l c c c}
\toprule
\multirow{2}{*}{\textbf{Dataset}} & \multirow{2}{*}{\textbf{Model}} & \multicolumn{3}{c}{\textbf{Overall Test}} \\ 
\cmidrule(lr){3-5} 
&  & \textbf{P} & \textbf{R} & \textbf{F1} \\
\midrule
\multirow{5}{*}{\parbox{1.5cm}{JNLPBA}}
& BERT-Large-Uncased  & 75.68 & 82.96 & 79.15 \\
& BERT-Large-Cased & 75.27 & 84.25 & 79.51 \\
& BiomedBERT-Large  & 76.75 & 84.60 & 80.49 \\
& DeBERTa-v3-Large  & 77.26 & 85.91 & 81.36 \\
& Mistral-7B  & 78.93 & 86.96 & \textbf{82.75} \\
& Llama-8B  & 78.00 & 86.65 & 82.10 \\
\midrule
\multirow{5}{*}{\parbox{1.5cm}{BioRED}}
& BERT-Large-Uncased & 86.91 & 90.43 & 88.64 \\
& BERT-Large-Cased & 85.88 & 91.87 & 88.77 \\
& BiomedBERT-Large  & 90.37 & 94.07 & 92.19 \\
& DeBERTa-v3-Large  & 91.37 & 94.15 & \textbf{92.74} \\
& Mistral-7B & 91.76 & 91.06 & 91.41 \\
& Llama-8B   & 91.33 & 90.58 & 90.96 \\
\midrule
\multirow{5}{*}{\parbox{1.5cm}{ChemProt}}
& BERT-Large-Uncased & 56.82 & 43.50 & 49.27 \\
& BERT-Large-Cased & 62.17 & 41.83 & 50.01 \\
& BiomedBERT-Large  & 74.50 & 43.91 & 55.26 \\
& DeBERTa-v3-Large & 56.22 & 46.39 & 50.83 \\
& Mistral-7B  & 78.35 & 43.22 & 55.71 \\
& Llama-8B  & 76.31 & 44.75 & \textbf{56.41} \\
\midrule
\multirow{5}{*}{\parbox{1.5cm}{BC5CDR}}
& BERT-Large-Uncased & 84.63 & 88.86 & 86.69 \\
& BERT-Large-Cased & 86.63 & 89.69 & 88.13 \\
& BiomedBERT-Large  & 86.99 & 92.28 & 89.56 \\
& DeBERTa-v3-Large  & 87.90 & 91.61 & 89.71 \\
& Mistral-7B & 91.04 & 91.67 & \textbf{91.35} \\
& Llama-8B  & 90.28 & 90.20 & 90.24 \\
\midrule
\multirow{5}{*}{\parbox{1.5cm}{Reddit-Impacts}}
& BERT-Large-Uncased  & 27.30 & 56.11 & 36.73 \\
& BERT-Large-Cased & 28.97 &  43.45 & 34.76 \\
& BiomedBERT-Large  & 32.84 & 40.36 & 36.22\\
& DeBERTa-v3-Large & 34.87 & 47.70 & 40.29 \\
& Mistral-7B & 44.92 & 33.12 & 38.13 \\
& Llama-8B   & 42.67 & 40.00 & \textbf{41.29} \\
\bottomrule
\end{tabular}%
\end{table}

While LLMs are scoring better over encoders in general, it is important to emphasize that the margin is small ($\approx 2\%$) for three of the datasets; for one dataset LLMs and encoders are almost equal. 
Considering this, it is important to also look at the average inference times for LLMs as shown in Table~\ref{tbl:Inference_AvgTimes}. For example, with the JNLPBA dataset, the best scoring encoder is 40 times faster on average than the best LLM (with a 2 point F-score difference). This pattern consistently applies to other datasets as well. Decoders' sequential generation of output tokens combined with the need to enumerate all input tokens (even if most of them are outside O tags) is much slower than the parallel simultaneous token classification in encoders.
At this point, we recall that two H100s were needed for LLM inference while only one was sufficient for encoder inference. So LLMs can also be cost prohibitive on top of being much slower than encoders. With these trade-offs, for domains where small performance differences are tolerable, encoders are still the better choice. 

Based on reviewer feedback, we also wanted to check if the observed performance trends as per strict measures hold in the relaxed setting which allows for partial matches. Especially for two- and three-word entities, relaxed measure based evaluation may not necessarily be bad, especially if the head word is matched. For the gold span ``breast cancer'' (where the head word is ``cancer''), missing the first word is not obviously ideal, but if the second word is captured, at least we are still capturing the essence of the entity (that it is a cancer) to some extent. We calculated partial match-based relaxed scores as shown in Table~\ref{tbl:Relaxed_Precision_Recall_F1}. Here, considering longer entities separately is not interesting, given we are measuring partial matches. As the table shows, the high level trends remain the same in overall scores. Even here, except for BioRED, LLMs are better in general, but only by two points maximum. The only exception to this small gap, under the stricter lens, was the Reddit-Impacts dataset, for which LLMs had an eight point gain. Even this does not exist anymore as that gap is shrunk to just a point in overall F-score. This is not  surprising since a match of a single token implies success even for longer entities. The SemEval relaxed measure we used does not impose the constraint that the partial match must include the head word of the full span. Hence, we believe readers should rely more on the strict match results in Table~\ref{tbl:Results}.

\section{Caveats}
All our findings are based on a large number of experiments and our implications are sound. However, it is important to discuss some caveats. To begin with, our findings are limited to English datasets and additional experiments are needed to confirm them for non-English languages. 
Hyperparameter tuning was straightforward and computationally less expensive to conduct for encoders. 
Tuning the learning rate and batch size did improve their performance. However, similar fine grained sweep of all hyperparameters was not viable for LLMs since each such configuration would take up multiple hours or days.  Across five datasets and two different LLMs, this was not tractable with the resources we had. Likewise, we set the QLoRA trainable component of LLMs to be roughly equal to the size of the encoder models used. An even bigger trainable QLoRA component may have garnered more performance gains. However, we report that using the full 7B/8B LLMs without any QLoRA adaptation made the performances worse. This could be potentially due to extreme overfitting to the training data with a very large parameter space.  


The LLM average inference times for BioRED (in Table~\ref{tbl:Inference_AvgTimes}) indicate that the best encoder (BiomedBERT) was nearly 220 times faster than the best decoder (Llama model). This is the only dataset where encoders are two orders of magnitude faster than LLMs. Some of this is attributable to the fact that the average input size of BioRED test instances is over 100 tokens, while all other datasets are in the 26 token range. Since BIO tagging necessitates enumeration of all tokens with a tag (even if it simply the `O' tag), BioRED naturally takes much longer with LLMs.  
However, we found another reason involving entity tag names. The entity types in BioRED had long names such as ``DiseaseOrPhenotypicFeature'' and we created BIO tags with these long names prefixed by ``B/I/O-''. Since these are not coded as special tokens, LLMs split these tags into multiple tokens adding to the overhead of generating multiple tokens per tag. While an order of magnitude more inference time is virtually guaranteed (due to longer input size), this 220X cost for BioRED is unlikely if tags are coded as special tokens.

Alternative approaches such as UniversalNER \cite{zhouuniversalner}, which was not explored in our study, offer a knowledge-distillation technique that transfers LLM capabilities into smaller, task-specific models. Since UniversalNER has demonstrated strong performance in general NER tasks, future work should evaluate whether it provides a computationally efficient alternative to the LLMs tested in our study.

\section{Conclusion}
In this effort, we conducted a series experiments to assess the potential of decoder LLMs in surpassing encoder models for biomedical NER. Our systematic evaluation revealed that LLMs appear to be consistently better than encoder models. However, depending on the dataset, the gains may be marginal at times but could also be substantial when long multi-token entities are involved. Other involved factors are prohibitive cost of hardware to run LLMs and high inference time, which is at least an order of magnitude more than that for encoder models. This tilts the scale in favor of encoder models when ample training data is available in certain domains, especially when real time interactive systems are needed. On the other hand, when marginal gains  matter (for high stakes decision making) or when gains are substantial and high costs are tolerable, one may choose LLMs as they appear to be getting better with time. Another unambiguous finding is that even when encoders come close to LLMs in overall performance, for longer entities, LLMs appear to be clear winners. So if cost is not a factor, there could be a way to combine encoder models and LLMs in an ensemble setup for further gains. 
As examples, the UL2 architecture\cite{tayul2} and
the more recent GPT-BERT formulation~\cite{charpentier2024bert} could be used in future assessments. 


\section*{Acknowledgement}
This work is supported by the U.S.~NIH  National Institute on Drug Abuse through grant R01DA057686 and the National Library of Medicine through grant R01LM013240.
The content is solely the responsibility of the authors and does not necessarily represent the official
views of the NIH.

\bibliographystyle{IEEEtran}
\bibliography{references}

\end{document}